\documentclass[11pt]{article}
\usepackage{coling2020}
\usepackage{times}
\usepackage{url}
\usepackage{latexsym}

\usepackage{multirow}
\usepackage{graphics}
\usepackage{adjustbox}
\usepackage{placeins}
\usepackage{float}
\restylefloat{table}

\colingfinalcopy 

\title{LIIR at SemEval-2020 Task 12:  A Cross-Lingual Augmentation Approach for Multilingual Offensive Language Identification}

\author{Erfan Ghadery \\
  Department of Computer Science \\
  KU Leuven\\
  {\ erfan.ghadery@student.kuleuven.be} \\\And
  Marie-Francine Moens \\
  Department of Computer Science \\
  KU Leuven \\
  {\ Sien.Moens@cs.kuleuven.be} \\}

\date{}

\begin{document}
\maketitle
\begin{abstract}
This paper presents our system entitled `LIIR' for SemEval-2020 Task 12 on Multilingual Offensive Language Identification in Social Media (OffensEval 2). We have participated in Subtask A for English, Danish, Greek, Arabic, and Turkish languages. We adapt and fine-tune the BERT and multilingual Bert models made available by Google AI\footnote{https://github.com/google-research/bert} for English and non-English languages respectively. For the English language, we use a combination of two fine-tuned BERT models. For other languages, we propose a cross-lingual augmentation approach in order to enrich training data and we use multilingual BERT to obtain sentence representations. LIIR achieved rank 14/38, 18/47, 24/86, 24/54, and 25/40 in Greek, Turkish, English, Arabic, and Danish languages, respectively.
\end{abstract}

\section{Introduction}
\label{intro}
\blfootnote{
    \hspace{-0.65cm}  
    This work is licensed under a Creative Commons 
    Attribution 4.0 International Licence.
    Licence details:
    \url{http://creativecommons.org/licenses/by/4.0/}.}

Nowadays, with an exponential increase in the use of social media platforms such as Facebook and Twitter by people from different educational and cultural backgrounds, the need for automatic methods for recognizing and filtering offensive languages is necessary \cite{chen2012detecting,nobata2016abusive}. Different types of offensive content like hate speech \cite{malmasi2018challenges}, aggression \cite{kumar2018benchmarking} and cyberbullying \cite{dinakar2011modeling} can be very harmful to the user's mental health, especially to children and youth \cite{xu2010filtering}.
\par The OffensEval 2019 competition \cite{zampieri2019semeval} was an attempt to build systems capable of recognizing offensive content in
social networks for the English language. The OffensEval 2019 organizers defined three Subtasks: whether a message is offensive or not (Subtask A), what is the type of the offensive message (Subtask B), and who is the target of the offensive message (Subtask C). This year, they have extended the competition to several languages while the Subtasks remain the same as in OffensEval 2019. OffensEval 2020 \cite{zampieri-etal-2020-semeval} features a multilingual dataset with five languages including English, Danish, Turkish, Greek, and Arabic.

\par This article presents our approaches to SemEval-2020 Task 12: OffensEval 2 - Multilingual Offensive Language Identification in Social Media. We have participated in Subtask A for all languages. The goal of Subtask A is recognizing if a sentence is offensive or not. For the English language, we separately fine-tune two bidirectional encoder representations of the BERT transformer architecture \cite{devlin2018bert} on two different datasets, and we use the combination of these two models for training our classifier. Also, we perform an extensive preprocessing for the English language. For other languages, we enhance the provided training dataset for each language using a cross-lingual augmentation approach, then, we train a classifier by fine-tuning a multilingual BERT (mBERT) \cite{devlin2018bert} using the augmented dataset with a linear classification layer on top. Our proposed augmentation approach, inspired by the works of \cite{lample2019cross,singh2019xlda}, translates each training sample into three other languages, then, adds the original training sample concatenated with every translation to the training set. LIIR has achieved a competitive performance with the best methods proposed by participants in the competition for the English track. Also, empirical results show that our cross-lingual augmentation approach is effective in improving results.

\par The rest of the article is organized as follows. The next section reviews related works. Section 3 describes the methodology of our proposed models. We will discuss experiments in Section 4 and the results are presented in Section 5.  Finally, the last section contains the conclusion of our work.

\section{Related Work}
\subsection{Offensive Language Identification}
Earlier works for addressing Offensive Language Identification relied on manually extracting the different types of features \cite{schmidt2017survey} such as token and character n-grams, word clusters, sentiment analysis outcomes, lexical and linguistic features, knowledge-based features, and multimodal information \cite{mehdad2016characters,warner2012detecting,gitari2015lexicon,dinakar2012common,hosseinmardi2015detection}. The extracted features were used to train machine learning methods like a support vector machine (SVM), naive Bayes, logistic regression, random forest classifier, or a neural network.
\par With the success of transfer learning enabled by pre-trained language models such as BERT \cite{devlin2018bert}, GPT \cite{radford2018improving}, and ULMFiT \cite{howard2018universal}, researchers have resorted to using these methods for addressing the Offensive Language Identification task. In the OffensEval 2019 competition \cite{zampieri2019semeval}, among the top-10 teams participated in Subtask A, seven used BERT with variations in the parameter settings and the preprocessing steps \cite{liu2019nuli,nikolov2019nikolov,pelicon2019embeddia}.

\subsection{Multilingual Methods}
There is a substantial body of work that investigates how to leverage multilingual data to improve the performance of a monolingual model or even to enable zero-shot classification. XLM model \cite{lample2019cross} extended BERT to a cross-lingual setting in which instead of monolingual text, it used concatenated parallel sentences in the pretraining procedure. This method achieved strong results in machine translation, language modeling, and cross-lingual Natural Language Inference. XLDA \cite{singh2019xlda} is a cross-lingual data augmentation method that simply replaces a segment of the input text with its translation in another language. The authors observed that most languages are effective as cross-lingual
augmenters in cross-lingual Natural Language Inference and Question Answering tasks. MNCN model \cite{ghadery2019mncn} utilized multilingual word embeddings as word representations and augmented training data by combining training sets in different languages for aspect-based sentiment analysis. This method is capable of classifying sentences in a specific language when there is no labeled training data available.

\section{Methodology}
In this section, we present the proposed methods in more detail. We have participated in Subtask A - categorizing a given sentence as `Offensive' or `Not-offensive' - for the English, Turkish, Arabic, Danish, and Greek languages. This year, OffensEval organizers have provided labeled training data for all the languages except for English where they just have provided unlabeled training data. Therefore, we propose two different approaches in this paper, one for the English language, and one for the other languages. For the English language, we fine-tune two BERT models separately on two different datasets, and we use the combination of these two models in training our classifier. For the other languages, a cross-lingual augmentation approach is used for enriching each language's training set, and we fine-tune an mBERT model to obtain sentence representations. In the following subsections, we describe our cross-lingual augmentation technique, and detail the proposed models.

\subsection{Cross-lingual Augmentation}
Given X = $\{x_{\ell},y_{\ell}\}_{\ell = 1}^{n}$ as a training set, where \begin{math}x \end{math} is a training sentence and \begin{math}y \end{math} is the corresponding label to \begin{math}x \end{math}, and \begin{math}n \end{math} is the number of train sentences, we create the augmented training set $\hat{X}$ in three steps as follows. First, $x_{\ell}$ is translated to English, French, and German languages using Google Translate \footnote{https://translate.google.com/}. In the second step, given the obtained translations $x_{{\ell}-{en}}$, $x_{{\ell}-{fr}}$, and $x_{{\ell}-{de}}$ as English, French, and German translations respectively, we generate three new samples as follows:
\par $\hat{x}_{\ell1}$ = $\{x_{\ell};x_{{\ell}-{en}},y_{\ell}\}$
\par $\hat{x}_{\ell2}$ = $\{x_{\ell};x_{{\ell}-{fr}},y_{\ell}\}$
\par $\hat{x}_{\ell3}$ = $\{x_{\ell};x_{{\ell}-{de}},y_{\ell}\}$
\par where ; is the concatenation operand. Finally, we create the augmented training set $\hat{X}$ by adding the original training samples and their three generated samples to $\hat{X}$. Choosing these three languages as translation candidates is because they are the top three languages used in Wikipedia. Since we know that the mBERT model is trained on a huge amount of Wikipedia page texts in different languages, by translating each training sample into the top three languages in this fine-tuning procedure, we make the representation of a sentence more informative. In other words, to predict the target label, the model can leverage the translated context if the original context is not sufficient \cite{lample2019cross}. It is fair to say that the proposed cross-lingual augmentation quality depends on the quality of the translation. 

\subsection{Models}
\subsubsection{English}

 For the English language, first of all, we automatically label the provided unlabeled dataset to obtain a weakly labeled training set. The OffensEval organizers have provided a confidence score for each sentence instead of a gold label, where the scores are the average confidence of belonging to the `Offensive' class produced by several learning methods. We investigated different threshold values for using the confidence scores for weakly labeling the sentences as `Offensive' or `Not-offensive' samples. In our experiments, we realized that the precise determination of a threshold for the confidence score is not an important factor in the performance but the important factor is the number of weakly labeled training samples. In order to decrease noise samples and since precision is a more important factor than recall in acquiring true training samples, we label sentences with confidence score more than 0.8 as `Offensive' and sentences with confidence score less than 0.2 as `Not-offensive'. Then, we randomly sample 300k `Offensive' sentences and 300k `Not-offensive' sentences as our final weakly labeled dataset. In the next step, we adapt and fine-tune two separate BERT models on the Offensive Language Identification dataset (OLID) \cite{zampierietal2019OLID} and our weakly labeled offensive dataset. Then, we train a feed-forward layer to classify a given sentence as `Offensive' or `Not-Offensive,' while the input of the classifier is the concatenation of the sentence representations extracted from the two fine-tuned BERT models. The representation of 'CLS' token, which is is first token of every input sequence, is considered as the sentence representation. 

\subsubsection{Other Languages}
\par For other languages, we augment the training set using the proposed cross-lingual augmentation technique, then, using the augmented dataset, we train a classifier by fine-tuning a pre-trained mBERT model topped with a feed-forward classification layer. The `CLS' token representation is fed to the classification layer as the sentence representation.

\section{Experiments}
\subsection{Datasets}
In this section, we present an overview of the
datasets used in this article for training our models for the OffensEval-2020 competition. For the English language, we use the large unlabeled dataset provided by the organizers \cite{rosenthal2020} to create the weakly labeled dataset. Also, we use the OLID \cite{zampierietal2019OLID} dataset for training the English model. For other languages, we utilize the provided labeled datasets by the organizers for Turkish \cite{coltekikin2020}, Danish \cite{sigurbergsson2020offensive}, Greek \cite{pitenis2020}, and Arabic \cite{mubarak2020arabic} languages.The detailed statistics of the datasets are summarized in Table 1. 

\begin{table}[H]
\centering
\begin{adjustbox}{width=220pt,center}
\begin{tabular}{|c|c|c|c|c|c|c|}
\hline
\multicolumn{1}{|c|}{\multirow{2}{*}{\textbf{Language}}} & \multicolumn{3}{c|}{\textbf{Train}} & \multicolumn{3}{c|}{\textbf{Test}} \\ \cline{2-7} 
\multicolumn{1}{|c|}{}                          
& \textbf{OFF}    & \textbf{NOT}    & \textbf{Total}    & \textbf{OFF}    & \textbf{NOT}    & \textbf{Total}   \\ \hline
English
& 300k & 300k & 600k & 1080 & 2807 & 3887 \\ \hline
Danish
& 307 & 2061 & 2368 & 41 & 288 & 329  \\ \hline
Turkish
& 4837 & 20184 & 25021 & 716 & 2812 & 3528  \\ \hline
Arabic
& 1371 & 5468 & 6839 & 402 & 1598 & 2000  \\ \hline
Greek
& 1989 & 5005 & 6994 & 242 & 1302 & 1544  \\ \hline
\end{tabular}
\end{adjustbox}
\caption{Datasets statistics}
\end{table}

\subsection{Experimental Settings}
For the English language, an extensive preprocessing is conducted including emoji to text projection \footnote{https://github.com/carpedm20/emoji}, hashtag segmentation \footnote{https://github.com/grantjenks/python-wordsegment}, replacing slang and abbreviations \cite{effrosynidis2017comparison}, replacing @USER by $<$user$>$ and `URL' by `http', and removing numbers. As the evaluation set, we held out 20 percent of the training set for Danish, Greek, and Turkish. For the Arabic language, the evaluation set is provided by the organizers and for the English language, the OffensEval-2019 test data is utilized as the evaluation set. The HuggingFaces library \cite{Wolf2019HuggingFacesTS} is used for obtaining pre-trained BERT and mBERT models. All the hyperparameters of our models are tuned using the validation data via grid search. We trained our models for 4 epochs with batch sizes of 8, 16, 24, 32, and 16 for English, Danish, Arabic, Greek, and Turkish, respectively. Adam optimizer is used with learning rates of 2e-05, 1e-05, 3e-05, 2e-05, and 2e-05 for English, Danish, Arabic, Greek, and Turkish, respectively. 
    
\section{Evaluation and Results}
\subsection{Results}
In this section, we present the results obtained by our methods on the test sets for Subtask A. Table 2 shows the results obtained by the submitted final models for each language on the test sets. All results are provided in terms of macro-F1. Furthermore, we provide the results obtained by two baselines, the `Majority' baseline, and the `Best system' baseline, for the sake of comparison. In the `Majority' baseline, the classifier simply predicts the majority class, and for the `Best system' baseline, we report the best result obtained by participant teams for each language. The best result for each language is marked in bold. Results show that LIIR has achieved a competitive performance compared to the best results obtained by the best teams in almost all the languages. These results demonstrate that our models can effectively identify offensive content contained in a given tweet. Also, we can observe that our method shows a weak performance in the Danish language which we believe is because of the problem of the mBERT model in processing some Danish characters \cite{stromberg2020danish}.

\begin{table}[H]
\label{main_res}
\centering
\begin{tabular}{|l|l|l|l|l|l|}
\hline
\textbf{System}  & \textbf{Turkish} & \textbf{Arabic} & \textbf{Greek}  & \textbf{Danish} & \textbf{English}\\ \hline
Majority baseline & 0.4435  & 0.4441 & 0.4575 & 0.4668 & 0.4193\\ \hline
Best system       & \textbf{0.8258}  & \textbf{0.9017} & \textbf{0.8522} & \textbf{0.8119} & \textbf{0.9204}\\ \hline
LIIR        & 0.7720  & 0.8418 & 0.8148 & 0.7019 & 0.9103\\ \hline
\end{tabular}
\caption{The macro-F1 scores obtained by LIIR compared to the baselines on the Test sets for each language for Subtask A.}
\end{table}

\subsection{Ablation Analysis}
In this part, we provide an ablation study on the models proposed for the different languages on the validation set. We show the effect of using two pre-trained models on two different datasets for classifying the English tweets. Furthermore, we examine how well the final results of other languages were influenced by the cross-lingual augmentation technique. Table 3 shows the ablation study results of the English language. The first observation is that the pre-trained model on the OLID dataset contributes to better performance compared to the pre-trained model on our weakly labeled dataset. The result is what we expected since the weakly labeled dataset unavoidably contains noise samples that will negatively affect model performance. The best result is obtained by a combination of the two models in the training procedure. In Table 4, the effect of the cross-lingual augmentation technique is shown. As the results show, the cross-lingual augmentation approach is quite effective in improving the model performance for all languages.

\begin{table}[H]
\centering
\begin{tabular}{|l|c|c|}
\hline
\textbf{System}              & \textbf{Macro-F1} & \textbf{Accuracy} \\ \hline
LIIIR \hspace{0.3cm} -- OLID        &     0.7463   &  0.7721        \\ \hline
LIIIR \hspace{0.3cm} -- Weak Datast &     0.7991   &  0.8477    \\ \hline
LIIR             & \textbf{0.8239} & \textbf{0.8628}          \\ \hline
\end{tabular}
\caption{Ablation analysis for the English language on the validation set.}
\end{table}

\begin{table}[H]
\centering
\begin{adjustbox}{width=450pt,center}
\centering
\begin{tabular}{|l|c|c|c|c|c|c|c|c|}
\hline
\multirow{2}{*}{\textbf{System}} & \multicolumn{2}{c|}{\textbf{Danish}} & \multicolumn{2}{c|}{\textbf{Turkish}} & \multicolumn{2}{c|}{\textbf{Arabic}} & \multicolumn{2}{c|}{\textbf{Greek}} \\ \cline{2-9} 
                        & \textbf{Macro-F1}     & \textbf{Accuracy}     & \textbf{Macro-F1}      &\textbf{Accuracy}     & \textbf{Macro-F1}     &\textbf{Accuracy}     & \textbf{Macro-F1}     & \textbf{Accuracy}    \\ \hline
LIIIR \hspace{0.3cm} -- Augmentation        & 0.8240 & 0.9307 & 0.7597 & 0.8684 & 0.8561 & 0.9180 & 0.7987 & 0.8422            \\ \hline
LIIIR \hspace{0.3cm} + Augmentation       & \textbf{0.8401} & \textbf{0.9358} & \textbf{0.7775} & \textbf{0.8734} & \textbf{0.8706} & \textbf{0.9280} & \textbf{0.8153} & \textbf{0.8565}            \\ \hline
\end{tabular}
\end{adjustbox}
\caption{Ablation analysis for the Danish, Turkish, Arabic, and Greek languages on the validation set.}
\end{table}

\section{Conclusion}
In this paper, we have presented our models for recognizing offensive content in the SemEval-2020 task 12 Subtask A for all languages. We have fine-tuned two BERT models on two different datasets for the English language. Moreover, we have fine-tuned an mBERT model on the augmented training sets for the other languages by implementing a cross-lingual augmentation approach. The evaluation results show that the proposed systems are capable of effectively recognizing offensive content in language. As future work, we intend to investigate other augmentation techniques. Furthermore, we plan to address the problem of imbalance in the training set.

\bibliographystyle{coling}
\bibliography{coling2020}

\begin{thebibliography}{}

\bibitem[\protect\citename{\c{C}\"{o}ltekin}2020]{coltekikin2020}
\c{C}a\u{g}r{\i} \c{C}\"{o}ltekin.
\newblock 2020.
\newblock {A Corpus of Turkish Offensive Language on Social Media}.
\newblock In {\em Proceedings of the 12th International Conference on Language
  Resources and Evaluation}. ELRA.

\bibitem[\protect\citename{Chen \bgroup et al.\egroup }2012]{chen2012detecting}
Ying Chen, Yilu Zhou, Sencun Zhu, and Heng Xu.
\newblock 2012.
\newblock Detecting offensive language in social media to protect adolescent
  online safety.
\newblock In {\em 2012 International Conference on Privacy, Security, Risk and
  Trust and 2012 International Confernece on Social Computing}, pages 71--80.
  IEEE.

\bibitem[\protect\citename{Devlin \bgroup et al.\egroup }2018]{devlin2018bert}
Jacob Devlin, Ming-Wei Chang, Kenton Lee, and Kristina Toutanova.
\newblock 2018.
\newblock Bert: Pre-training of deep bidirectional transformers for language
  understanding.
\newblock {\em arXiv preprint arXiv:1810.04805}.

\bibitem[\protect\citename{Dinakar \bgroup et al.\egroup
  }2011]{dinakar2011modeling}
Karthik Dinakar, Roi Reichart, and Henry Lieberman.
\newblock 2011.
\newblock Modeling the detection of textual cyberbullying.
\newblock In {\em fifth international AAAI conference on weblogs and social
  media}.

\bibitem[\protect\citename{Dinakar \bgroup et al.\egroup
  }2012]{dinakar2012common}
Karthik Dinakar, Birago Jones, Catherine Havasi, Henry Lieberman, and Rosalind
  Picard.
\newblock 2012.
\newblock Common sense reasoning for detection, prevention, and mitigation of
  cyberbullying.
\newblock {\em ACM Transactions on Interactive Intelligent Systems (TiiS)},
  2(3):1--30.

\bibitem[\protect\citename{Effrosynidis \bgroup et al.\egroup
  }2017]{effrosynidis2017comparison}
Dimitrios Effrosynidis, Symeon Symeonidis, and Avi Arampatzis.
\newblock 2017.
\newblock A comparison of pre-processing techniques for twitter sentiment
  analysis.
\newblock In {\em International Conference on Theory and Practice of Digital
  Libraries}, pages 394--406. Springer.

\bibitem[\protect\citename{Ghadery \bgroup et al.\egroup
  }2019]{ghadery2019mncn}
Erfan Ghadery, Sajad Movahedi, Heshaam Faili, and Azadeh Shakery.
\newblock 2019.
\newblock Mncn: A multilingual ngram-based convolutional network for aspect
  category detection in online reviews.
\newblock In {\em Proceedings of the AAAI Conference on Artificial
  Intelligence}, volume~33, pages 6441--6448.

\bibitem[\protect\citename{Gitari \bgroup et al.\egroup
  }2015]{gitari2015lexicon}
Njagi~Dennis Gitari, Zhang Zuping, Hanyurwimfura Damien, and Jun Long.
\newblock 2015.
\newblock A lexicon-based approach for hate speech detection.
\newblock {\em International Journal of Multimedia and Ubiquitous Engineering},
  10(4):215--230.

\bibitem[\protect\citename{Hosseinmardi \bgroup et al.\egroup
  }2015]{hosseinmardi2015detection}
Homa Hosseinmardi, Sabrina~Arredondo Mattson, Rahat~Ibn Rafiq, Richard Han, Qin
  Lv, and Shivakant Mishra.
\newblock 2015.
\newblock Detection of cyberbullying incidents on the instagram social network.
\newblock {\em arXiv preprint arXiv:1503.03909}.

\bibitem[\protect\citename{Howard and Ruder}2018]{howard2018universal}
Jeremy Howard and Sebastian Ruder.
\newblock 2018.
\newblock Universal language model fine-tuning for text classification.
\newblock {\em arXiv preprint arXiv:1801.06146}.

\bibitem[\protect\citename{Kumar \bgroup et al.\egroup
  }2018]{kumar2018benchmarking}
Ritesh Kumar, Atul~Kr Ojha, Shervin Malmasi, and Marcos Zampieri.
\newblock 2018.
\newblock Benchmarking aggression identification in social media.
\newblock In {\em Proceedings of the First Workshop on Trolling, Aggression and
  Cyberbullying (TRAC-2018)}, pages 1--11.

\bibitem[\protect\citename{Lample and Conneau}2019]{lample2019cross}
Guillaume Lample and Alexis Conneau.
\newblock 2019.
\newblock Cross-lingual language model pretraining.
\newblock {\em arXiv preprint arXiv:1901.07291}.

\bibitem[\protect\citename{Liu \bgroup et al.\egroup }2019]{liu2019nuli}
Ping Liu, Wen Li, and Liang Zou.
\newblock 2019.
\newblock Nuli at semeval-2019 task 6: Transfer learning for offensive language
  detection using bidirectional transformers.
\newblock In {\em Proceedings of the 13th International Workshop on Semantic
  Evaluation}, pages 87--91.

\bibitem[\protect\citename{Malmasi and Zampieri}2018]{malmasi2018challenges}
Shervin Malmasi and Marcos Zampieri.
\newblock 2018.
\newblock Challenges in discriminating profanity from hate speech.
\newblock {\em Journal of Experimental \& Theoretical Artificial Intelligence},
  30(2):187--202.

\bibitem[\protect\citename{Mehdad and Tetreault}2016]{mehdad2016characters}
Yashar Mehdad and Joel Tetreault.
\newblock 2016.
\newblock Do characters abuse more than words?
\newblock In {\em Proceedings of the 17th Annual Meeting of the Special
  Interest Group on Discourse and Dialogue}, pages 299--303.

\bibitem[\protect\citename{Mubarak \bgroup et al.\egroup
  }2020]{mubarak2020arabic}
Hamdy Mubarak, Ammar Rashed, Kareem Darwish, Younes Samih, and Ahmed Abdelali.
\newblock 2020.
\newblock Arabic offensive language on twitter: Analysis and experiments.
\newblock {\em arXiv preprint arXiv:2004.02192}.

\bibitem[\protect\citename{Nikolov and Radivchev}2019]{nikolov2019nikolov}
Alex Nikolov and Victor Radivchev.
\newblock 2019.
\newblock Nikolov-radivchev at semeval-2019 task 6: Offensive tweet
  classification with bert and ensembles.
\newblock In {\em Proceedings of the 13th International Workshop on Semantic
  Evaluation}, pages 691--695.

\bibitem[\protect\citename{Nobata \bgroup et al.\egroup
  }2016]{nobata2016abusive}
Chikashi Nobata, Joel Tetreault, Achint Thomas, Yashar Mehdad, and Yi~Chang.
\newblock 2016.
\newblock Abusive language detection in online user content.
\newblock In {\em Proceedings of the 25th international conference on world
  wide web}, pages 145--153.

\bibitem[\protect\citename{Pelicon \bgroup et al.\egroup
  }2019]{pelicon2019embeddia}
Andra{\v{z}} Pelicon, Matej Martinc, and Petra~Kralj Novak.
\newblock 2019.
\newblock Embeddia at semeval-2019 task 6: Detecting hate with neural network
  and transfer learning approaches.
\newblock In {\em Proceedings of the 13th International Workshop on Semantic
  Evaluation}, pages 604--610.

\bibitem[\protect\citename{Pitenis \bgroup et al.\egroup }2020]{pitenis2020}
Zeses Pitenis, Marcos Zampieri, and Tharindu Ranasinghe.
\newblock 2020.
\newblock {Offensive Language Identification in Greek}.
\newblock In {\em Proceedings of the 12th Language Resources and Evaluation
  Conference}. ELRA.

\bibitem[\protect\citename{Radford \bgroup et al.\egroup
  }2018]{radford2018improving}
Alec Radford, Karthik Narasimhan, Time Salimans, and Ilya Sutskever.
\newblock 2018.
\newblock Improving language understanding with unsupervised learning.
\newblock {\em Technical report, OpenAI}.

\bibitem[\protect\citename{Rosenthal \bgroup et al.\egroup
  }2020]{rosenthal2020}
Sara Rosenthal, Pepa Atanasova, Georgi Karadzhov, Marcos Zampieri, and Preslav
  Nakov.
\newblock 2020.
\newblock {A Large-Scale Weakly Supervised Dataset for Offensive Language
  Identification}.
\newblock In {\em arxiv}.

\bibitem[\protect\citename{Schmidt and Wiegand}2017]{schmidt2017survey}
Anna Schmidt and Michael Wiegand.
\newblock 2017.
\newblock A survey on hate speech detection using natural language processing.
\newblock In {\em Proceedings of the Fifth International Workshop on Natural
  Language Processing for Social Media}, pages 1--10.

\bibitem[\protect\citename{Sigurbergsson and
  Derczynski}2020]{sigurbergsson2020offensive}
Gudbjartur~Ingi Sigurbergsson and Leon Derczynski.
\newblock 2020.
\newblock {Offensive Language and Hate Speech Detection for Danish}.
\newblock In {\em Proceedings of the 12th Language Resources and Evaluation
  Conference}. ELRA.

\bibitem[\protect\citename{Singh \bgroup et al.\egroup }2019]{singh2019xlda}
Jasdeep Singh, Bryan McCann, Nitish~Shirish Keskar, Caiming Xiong, and Richard
  Socher.
\newblock 2019.
\newblock Xlda: Cross-lingual data augmentation for natural language inference
  and question answering.
\newblock {\em arXiv preprint arXiv:1905.11471}.

\bibitem[\protect\citename{Str{\o}mberg-Derczynski \bgroup et al.\egroup
  }2020]{stromberg2020danish}
Leon Str{\o}mberg-Derczynski, Rebekah Baglini, Morten~H Christiansen, Manuel~R
  Ciosici, Jacob~Aarup Dalsgaard, Riccardo Fusaroli, Peter~Juel Henrichsen,
  Rasmus Hvingelby, Andreas Kirkedal, Alex~Speed Kjeldsen, et~al.
\newblock 2020.
\newblock The danish gigaword project.
\newblock {\em arXiv preprint arXiv:2005.03521}.

\bibitem[\protect\citename{Warner and Hirschberg}2012]{warner2012detecting}
William Warner and Julia Hirschberg.
\newblock 2012.
\newblock Detecting hate speech on the world wide web.
\newblock In {\em Proceedings of the second workshop on language in social
  media}, pages 19--26. Association for Computational Linguistics.

\bibitem[\protect\citename{Wolf \bgroup et al.\egroup
  }2019]{Wolf2019HuggingFacesTS}
Thomas Wolf, Lysandre Debut, Victor Sanh, Julien Chaumond, Clement Delangue,
  Anthony Moi, Pierric Cistac, Tim Rault, R'emi Louf, Morgan Funtowicz, and
  Jamie Brew.
\newblock 2019.
\newblock Huggingface's transformers: State-of-the-art natural language
  processing.
\newblock {\em ArXiv}, abs/1910.03771.

\bibitem[\protect\citename{Xu and Zhu}2010]{xu2010filtering}
Zhi Xu and Sencun Zhu.
\newblock 2010.
\newblock Filtering offensive language in online communities using grammatical
  relations.
\newblock In {\em Proceedings of the Seventh Annual Collaboration, Electronic
  Messaging, Anti-Abuse and Spam Conference}, pages 1--10.

\bibitem[\protect\citename{Zampieri \bgroup et al.\egroup
  }2019a]{zampierietal2019OLID}
Marcos Zampieri, Shervin Malmasi, Preslav Nakov, Sara Rosenthal, Noura Farra,
  and Ritesh Kumar.
\newblock 2019a.
\newblock {Predicting the Type and Target of Offensive Posts in Social Media}.
\newblock In {\em Proceedings of NAACL}.

\bibitem[\protect\citename{Zampieri \bgroup et al.\egroup
  }2019b]{zampieri2019semeval}
Marcos Zampieri, Shervin Malmasi, Preslav Nakov, Sara Rosenthal, Noura Farra,
  and Ritesh Kumar.
\newblock 2019b.
\newblock Semeval-2019 task 6: Identifying and categorizing offensive language
  in social media (offenseval).
\newblock {\em arXiv preprint arXiv:1903.08983}.

\bibitem[\protect\citename{Zampieri \bgroup et al.\egroup
  }2020]{zampieri-etal-2020-semeval}
Marcos Zampieri, Preslav Nakov, Sara Rosenthal, Pepa Atanasova, Georgi
  Karadzhov, Hamdy Mubarak, Leon Derczynski, Zeses Pitenis, and
  \c{C}a\u{g}r{\i} \c{C}\"{o}ltekin.
\newblock 2020.
\newblock {SemEval-2020 Task 12: Multilingual Offensive Language Identification
  in Social Media (OffensEval 2020)}.
\newblock In {\em Proceedings of SemEval}.

\end{thebibliography}

\end{document}